\documentclass[conference]{IEEEtran}
\usepackage{cite}
\usepackage{amsmath,amssymb,amsfonts}
\usepackage{algorithmic}
\usepackage{graphicx}
\usepackage{textcomp}
\usepackage{xcolor}
\def\BibTeX{{\rm B\kern-.05em{\sc i\kern-.025em b}\kern-.08em
    T\kern-.1667em\lower.7ex\hbox{E}\kern-.125emX}}
\begin{document}

\title{Whither the Priors for (Vocal) Interactivity?}

\author{\IEEEauthorblockN{Roger K. Moore}
\IEEEauthorblockA{\textit{Dept. of Computer Science} \\
\textit{University of Sheffield}\\
Sheffield, UK \\
r.k.moore@sheffield.ac.uk}
}

\maketitle

\begin{abstract}
Voice-based communication is often cited as one of the most `natural' ways in which humans and robots might interact, and the recent availability of accurate automatic speech recognition and intelligible speech synthesis has enabled researchers to integrate advanced off-the-shelf spoken language technology components into their robot platforms.  Despite this, the resulting interactions are anything but `natural'.  It transpires that simply giving a robot a voice doesn't mean that a user will know how (or when) to talk to it, and the resulting `conversations' tend to be stilted, one-sided and short.  On the surface, these difficulties might appear to be fairly trivial consequences of users' unfamiliarity with robots (and \emph{vice versa}), and that any problems would be mitigated by long-term use by the human, coupled with `deep learning' by the robot.  However, it is argued here that such communication failures are indicative of a deeper malaise: a fundamental lack of basic principles -- \emph{priors} -- underpinning not only speech-based interaction in particular, but (vocal) interactivity in general.  This is evidenced not only by the fact that contemporary spoken language systems already require training data sets that are orders-of-magnitude greater than that experienced by a young child, but also by the lack of design principles for creating effective communicative human-robot interaction. This short position paper identifies some of the key areas where theoretical insights might help overcome these shortfalls.
\end{abstract}


\section{Introduction}

Recent years have seen considerable progress in the deployment of `intelligent' communicative agents such as Apple's \emph{Siri} and Amazon's \emph{Alexa} \cite{Boyd2018}.  However, effective speech-based human-robot dialogue is less well developed; not only do the fields of robotics and spoken language technology present their own special problems, but their combination raises an additional set of issues \cite{Moore2015,Marge2021}.  In particular, there appears to be a `habitability gap' between the formulaic behaviour that typifies contemporary spoken language dialogue systems and the rich and flexible nature of human-human conversation \cite{Moore2016x}.  It turns out that simply interfacing state-of-the-art speech technology with a state-of-the-art robot does not lead to effective human-robot interaction.

There are a number of reasons as to why this might be the case, and there are vibrant research communities dedicated to making continuous incremental improvements to the capabilities of such systems.  Such efforts are currently fuelled by the success of approaches based on `deep learning' using extraordinary quantities of (labelled) training data.  However, while `end-to-end' data-driven machine learning is remarkably effective, it is incredibly inefficient in comparison to the data requirements of a living system \cite{Marcus2019}.  For example, a contemporary large vocabulary continuous speech recognition system is trained on hundreds of thousands or even millions of hours of speech (e.g.\ \cite{Xiao2021}), whereas a linguistically-competent ten year-old child has been exposed to only around ten thousand hours of speech \cite{Moore2003a}.

Clearly, contemporary machine learning is missing some fundamental `priors', i.e.\ task-specific constraints that reflect the phylogenetic (as opposed to ontogenetic) trajectory of learning.  Hence, there are several areas in which \emph{theoretical} insights might offer not only improvements in usability and interaction, but also substantial gains in efficiency.

\section{Interactional Affordances}

\begin{figure}[htbp]
	\centerline{\includegraphics[width=0.9\columnwidth]{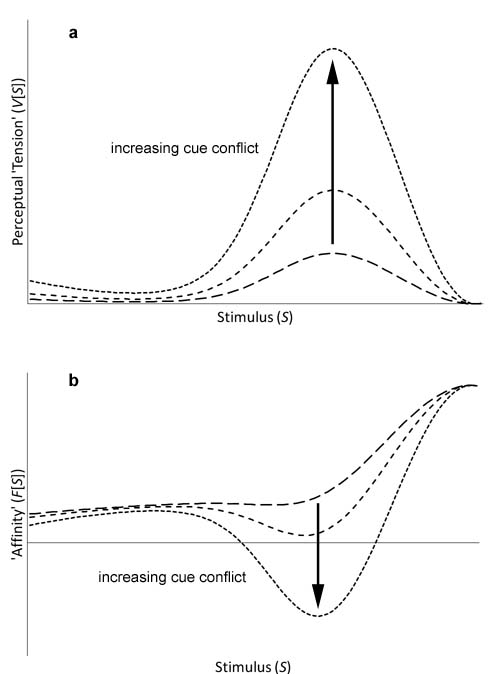}}
	\caption{Predicted output from the Bayesian model of the `uncanny valley' \cite{Moore2012} showing (a) an increase in perceptual tension leading to (b) a decrease in `affinity' as a consequence of conflicting perceptual cues (such as a robot with a mechanical face and a humanlike voice).}
	\label{fig:UV}
\end{figure}

First, \emph{robots are not people} \ldots\ and, as Mori \cite{Mori1970} pointed out many years ago, the more humanlike the robot, the more sensitive observers are to any inconsistencies in its make-up and behaviour.  Mori famously coined the term `uncanny valley' to describe the negative feelings evoked by near-human artefacts, and many empirical studies have confirmed the existence of the phenomenon \cite{MacDorman2009,Burleigh2013,MacDorman2015}, even for mismatched faces and voices \cite{Mitchell2011}.  However, it is only relatively recently that a mathematically-grounded (Bayesian) \emph{theory} has been proposed to explain the effect \cite{Moore2012} -- see Fig.~\ref{fig:UV}.

The Bayesian model associates the uncanny valley with the aforementioned habitability gap, and predicts that it can only be avoided if the visual, vocal, behavioural, interactional and cognitive \emph{affordances} of an artefact are aligned.  I.e. how a robot looks, sounds and behaves needs to be coherent.  Hence, given that the state-of-the-art in these areas varies, it implies that the capabilities of an artificial agent would be capped by the affordance with the \emph{lowest} performance -- and that will have a significant impact on the interactional complexity when it comes to the language component.  However, right now we lack the theoretical insights that are needed to extend the Bayesian model such that it is able to provide the necessary design criteria.

\section{Language and Communication}

Second, \emph{language is a uniquely human behaviour}, which means that it is only `natural' in a human-human context.  Almost nothing is known about language-based interaction between \emph{mismatched} interlocutors, and it has even been argued that it may even be an unsurmountable problem in a human-machine context \cite{Moore2016x}.  In particular, if spoken language interaction is viewed as being based on the co-evolution of two key traits -- \emph{ostensive-inferential} communication and \emph{recursive mind-reading} \cite{Scott-Phillips2015}, then contemporary voice-based systems are essentially only dealing with one aspect -- \emph{inference} -- and while some of the high-level issues relating to recursive mind-reading are being addressed \cite{Moore2017b}, we lack a theoretical framework relating to user engagement and interaction that would inform the implementation of practical mechanisms to handle not just what to say, but crucially when to say it and, in a group context, to whom.

\section{Cooperation, Coordination and Control}

Third, \emph{robots are complex collections of sensors, actuators, and control software that have to operate in real-time in real-world physical environments}.  This calls for the precise coordination of multiple `degrees-of-freedom' (DoF) of the robot itself, as well as orchestrating its interaction with its external environment which, of course, may itself contain other living/non-living agents.  Such complex control tasks are well beyond the reach of contemporary theoretical frameworks, and require a consolidation of many different concepts (especially rhythm \cite{Ravignani2014}, synchrony \cite{Kuramoto1975,Cummins2011,Strogatz2012} and dialogue \cite{Fusaroli2014}) into a single unified perspective.

One somewhat overlooked conceptual framework is `Perceptual Control Theory' (PCT) \cite{Powers1973}.  Rooted in `cybernetics' \cite{Wiener1948}, PCT provides a perspective based on multi-level closed-loop feedback-control mechanisms, and it has been shown to offer promise in modelling vocal interactivity \cite{Moore2019d}.  The main advantage of this approach is that the optimisation criteria can be made more explicit, thereby offering the potential to gain a deeper understanding of the implications of particular parameters/settings on the emergent collective behaviours

\begin{figure}[htbp]
	\centerline{\includegraphics[width=\columnwidth]{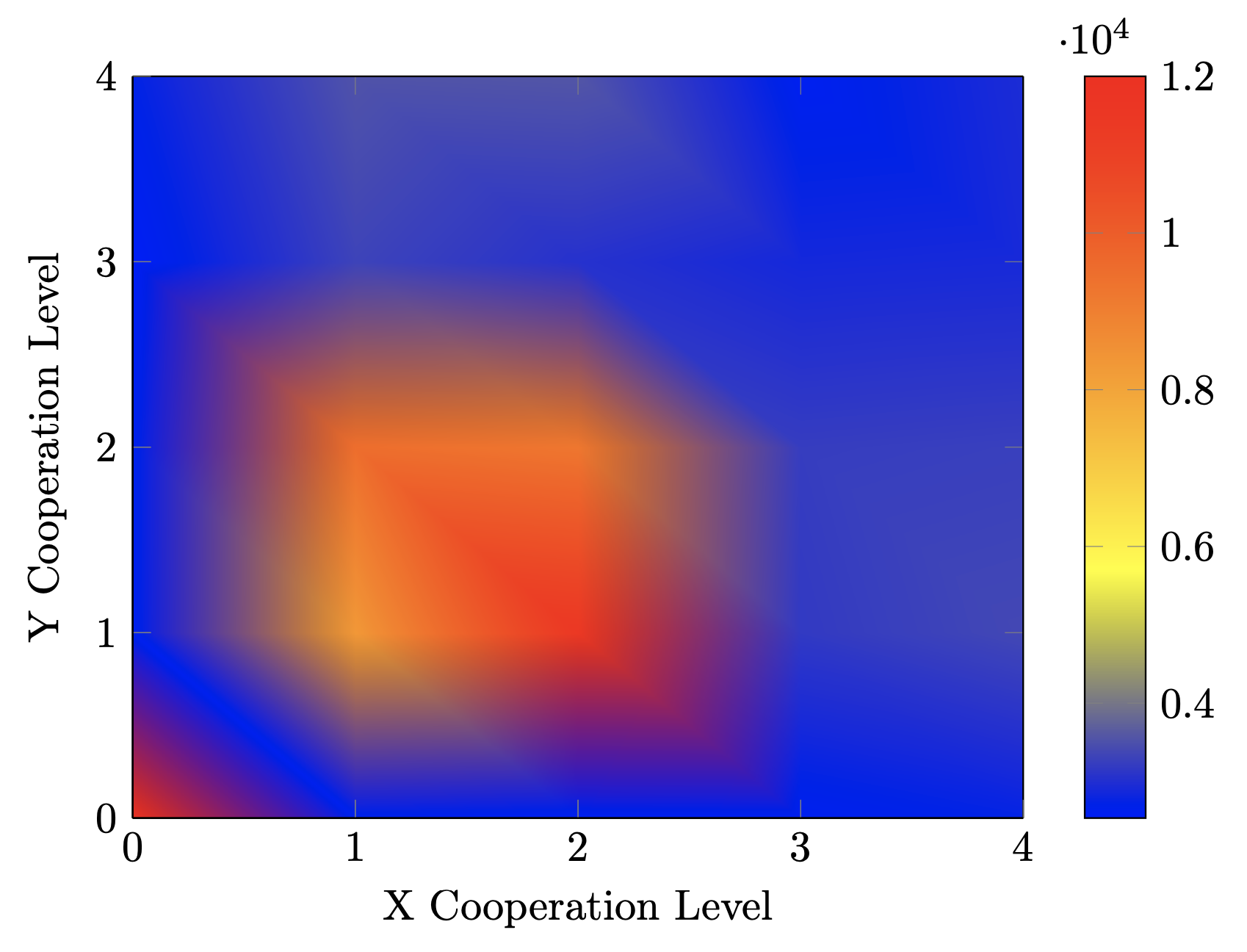}}
	\caption{Heat map depicting solutions times for a symmetric task undertaken by two (X and Y) agents with different levels of cooperation from each.  The results show that the task is not only solvable quickly with high levels of cooperation from both agents (top-right), but also with asymmetric levels of cooperation (left and bottom).}
	\label{fig:COOP}
\end{figure}

Such a framework is currently under investigation (by the author) in order to determine some general principles of cooperative behaviour between agents.  Preliminary (and, as yet, unpublished) results suggest that from an \emph{energetics} perspective, a key feature of successful cooperation may be the asymmetric allocation of responsibilities, even for a symmetric task -- see Fig.~\ref{fig:COOP}.

\section{Crossing Discipline Boundaries}

\begin{figure}[htbp]
	\centerline{\includegraphics[width=\columnwidth]{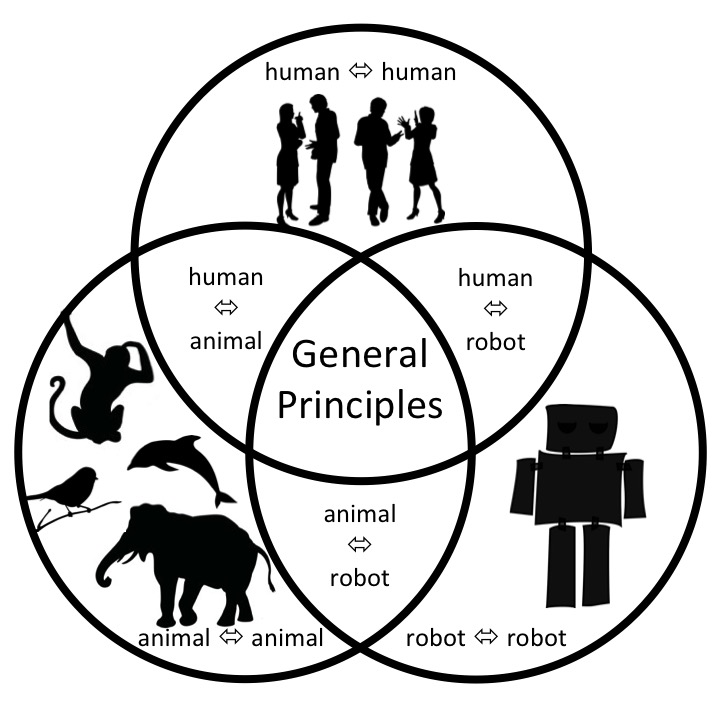}}
	\caption{Logo for the \emph{Vocal Interactivity in-and-between Humans, Animals and Robots} (VIHAR) series of international workshops with `general principles' at the core.}
	\label{fig:VIHAR}
\end{figure}

Finally, \emph{the challenges posed by voice-based human-robot interaction are inherently interdisciplinary}, yet the relevant research expertise typically exists in community-specific silos with little cross-fertilisation.  Therefore, in 2016 an attempt at forging a link between different communities was made with the publication of a position paper on `Vocal Interactivity in-and-between Humans, Animals and Robots' (VIHAR) \cite{Moore2016q} coupled with the organisation of a Dagstuhl Seminar on the same topic \cite{Moore2017g}.  The main aim was to create an opportunity for sharing different perspectives of vocal interactivity with a view to establishing some general principles -- see Fig.~\ref{fig:VIHAR}.

The Dagstuhl event led to the creation of a growing community of interest, with international workshops being held in 2017 (http://vihar-2017.vihar.org), 2019 (http://vihar-2019.vihar.org) and 2021 (http://vihar-2021.vihar.org) -- the latter attracting over 200 registered participants.  The published proceedings are all available on-line, and a special issue of \emph{Interaction Studies} is currently open for submissions (https://benjamins.com/catalog/is).

\section{Summary and Conclusion}

This short position paper has attempted to illustrate a range of different issues in (vocal) interactivity with a focus on the missing \emph{priors}, i.e.\ the fundamental principles on which (vocal) interactivity is based.  The key hypothesis is that contemporary approaches, which are now mostly based on deep learning, are grossly inefficient in terms of their use of data precisely because of the lack of a sound theoretical foundation for modelling (vocal) interaction.  Four areas of interest have been highlighted for attention: (i) interactional affordances, (ii) language and communication, (iii) cooperation, coordination and control, and (iv) crossing disciplinary boundaries.  However, the common theme is that the design of a non-living artefact (such as a robot) that is intended to interact with a living system (such as a human being) should be founded on an as yet unknown set of theoretical principles, if such interaction is to be efficient, effective and `habitable' by all of the living and non-living participants.

\bibliographystyle{IEEEtran}
\bibliography{IEEEabrv,mybibfile}

\begin{thebibliography}{10}
\providecommand{\url}[1]{#1}
\csname url@samestyle\endcsname
\providecommand{\newblock}{\relax}
\providecommand{\bibinfo}[2]{#2}
\providecommand{\BIBentrySTDinterwordspacing}{\spaceskip=0pt\relax}
\providecommand{\BIBentryALTinterwordstretchfactor}{4}
\providecommand{\BIBentryALTinterwordspacing}{\spaceskip=\fontdimen2\font plus
\BIBentryALTinterwordstretchfactor\fontdimen3\font minus
  \fontdimen4\font\relax}
\providecommand{\BIBforeignlanguage}[2]{{%
\expandafter\ifx\csname l@#1\endcsname\relax
\typeout{** WARNING: IEEEtran.bst: No hyphenation pattern has been}%
\typeout{** loaded for the language `#1'. Using the pattern for}%
\typeout{** the default language instead.}%
\else
\language=\csname l@#1\endcsname
\fi
#2}}
\providecommand{\BIBdecl}{\relax}
\BIBdecl

\bibitem{Boyd2018}
C.~Boyd, ``The past, present, and future of speech recognition technology,'' 1
  2018,
  https://medium.com/swlh/the-past-present-and-future-of-speech-recognition-technology-cf13c179aaf.

\bibitem{Moore2015}
R.~K. Moore, ``{From talking and listening robots to intelligent communicative
  machines},'' in \emph{Robots That Talk and Listen}, J.~Markowitz, Ed.\hskip
  1em plus 0.5em minus 0.4em\relax Boston, MA: De Gruyter, 2015, ch.~12, pp.
  317--335.

\bibitem{Marge2021}
M.~Marge, C.~Espy-Wilson, N.~G. Ward, A.~Alwan, Y.~Artzi, M.~Bansal,
  G.~Blankenship, J.~Chai, H.~D. III, D.~Dey, M.~Harper, T.~Howard,
  C.~Kennington, I.~Kruijff-Korbayov{\'{a}}, D.~Manocha, C.~Matuszek, R.~Mead,
  R.~Mooney, R.~K. Moore, M.~Ostendorf, H.~Pon-Barry, A.~I. Rudnicky,
  M.~Scheutz, R.~S. Amant, T.~Sun, S.~Tellex, D.~Traum, and Z.~Yu, ``{Spoken
  Language Interaction with Robots: Recommendations for Future Research},''
  \emph{Computer Speech and Language}, vol.~71, no. 101255, 2022.

\bibitem{Moore2016x}
R.~K. Moore, ``{Is spoken language all-or-nothing? Implications for future
  speech-based human-machine interaction},'' in \emph{Dialogues with Social
  Robots – Enablements, Analyses, and Evaluation}, K.~Jokinen and G.~Wilcock,
  Eds.\hskip 1em plus 0.5em minus 0.4em\relax Springer Lecture Notes in
  Electrical Engineering (LNEE), 2016.

\bibitem{Marcus2019}
G.~Marcus and E.~Davis, \emph{{Rebooting AI: Building Artificial Intelligence
  We Can Trust}}.\hskip 1em plus 0.5em minus 0.4em\relax Pantheon Books, 2019.

\bibitem{Xiao2021}
A.~Xiao, W.~Zheng, G.~Keren, D.~Le, F.~Zhang, C.~Fuegen, O.~Kalinli, Y.~Saraf,
  and A.~Mohamed, ``{Scaling ASR Improves zero and few shot learning},''
  \emph{arXiv}, 2021.

\bibitem{Moore2003a}
R.~K. Moore, ``{A comparison of the data requirements of automatic speech
  recognition systems and human listeners},'' in \emph{EUROSPEECH03}, Geneva,
  2003, pp. 2581--2584.

\bibitem{Mori1970}
M.~Mori, ``{Bukimi no tani (the uncanny valley)},'' \emph{Energy}, vol.~7, pp.
  33--35, 1970.

\bibitem{MacDorman2009}
K.~F. MacDorman, R.~D. Green, C.-C. Ho, and C.~Koch, ``{Too real for comfort:
  Uncanny responses to computer generated faces},'' \emph{Computers in Human
  Behavior}, vol.~25, pp. 695--710, 2009.

\bibitem{Burleigh2013}
T.~J. Burleigh, J.~R. Schoenherr, and G.~L. Lacroix, ``{Does the uncanny valley
  exist? An empirical test of the relationship between eeriness and the human
  likeness of digitally created faces},'' \emph{Computers in Human Behavior},
  vol.~29, no.~3, pp. 759--771, may 2013.

\bibitem{MacDorman2015}
K.~F. MacDorman and D.~Chattopadhyay, ``{Reducing consistency in human realism
  increases the uncanny valley effect; increasing category uncertainty does
  not.}'' \emph{Cognition}, vol. 146, pp. 190--205, oct 2015.

\bibitem{Mitchell2011}
W.~J. Mitchell, K.~A. {Szerszen Sr.}, A.~S. Lu, P.~W. Schermerhorn, M.~Scheutz,
  and K.~F. MacDorman, ``{A mismatch in the human realism of face and voice
  produces an uncanny valley},'' \emph{i-Perception}, vol.~2, no.~1, pp.
  10--12, 2011.

\bibitem{Moore2012}
R.~K. Moore, ``{A Bayesian explanation of the ‘Uncanny Valley' effect and
  related psychological phenomena},'' \emph{Nature Scientific Reports}, vol.~2,
  no. 864, 2012.

\bibitem{Scott-Phillips2015}
T.~Scott-Phillips, \emph{{Speaking Our Minds: Why human communication is
  different, and how language evolved to make it special}}.\hskip 1em plus
  0.5em minus 0.4em\relax London, New York: Palgrave MacMillan, 2015.

\bibitem{Moore2017b}
R.~K. Moore and M.~Nicolao, ``{Towards a Needs-Based Architecture for
  ‘Intelligent' Communicative Agents: Speaking with Intention},''
  \emph{Frontiers in Robotics and AI}, vol.~4, no.~66, 2017.

\bibitem{Ravignani2014}
A.~Ravignani, D.~L. Bowling, and W.~T. Fitch, ``{Chorusing, synchrony, and the
  evolutionary functions of rhythm},'' \emph{Frontiers in psychology}, vol.~5,
  p. 1118, jan 2014.

\bibitem{Kuramoto1975}
Y.~Kuramoto, ``{Self-entrainment of a population of coupled non-linear
  oscillators},'' in \emph{International Symposium on Mathematical Problems in
  Theoretical Physics}, H.~Araki, Ed., 1975, pp. 420--422.

\bibitem{Cummins2011}
F.~Cummins, ``{Periodic and aperiodic synchronization in skilled action},''
  \emph{Frontiers in Human Neuroscience}, vol.~5, no. 170, pp. 1--9, 2011.

\bibitem{Strogatz2012}
S.~H. Strogatz, \emph{{Sync: How Order Emerges from Chaos In the Universe,
  Nature, and Daily Life}}.\hskip 1em plus 0.5em minus 0.4em\relax Hatchette
  Book Group, 2012.

\bibitem{Fusaroli2014}
R.~Fusaroli, J.~R{\c{a}}czaszek-Leonardi, and K.~Tyl{\'{e}}n, ``{Dialog as
  interpersonal synergy},'' \emph{New Ideas in Psychology}, vol.~32, pp.
  147--157, jan 2014.

\bibitem{Powers1973}
W.~T. Powers, \emph{{Behavior: The Control of Perception}}.\hskip 1em plus
  0.5em minus 0.4em\relax NY: Aldine: Hawthorne, 1973.

\bibitem{Wiener1948}
N.~Wiener, \emph{{Cybernetics: or Control and Communication in the Animal and
  the Machine}}, 2nd~ed.\hskip 1em plus 0.5em minus 0.4em\relax Cambridge,
  Mass.: The MIT Press, 1965.

\bibitem{Moore2019d}
R.~K. Moore, ``{Vocal interactivity in crowds, flocks and swarms: implications
  for voice user interfaces},'' in \emph{2nd International Workshop on Vocal
  Interactivity in-and-between Humans, Animals and Robots (VIHAR-2019)},
  London, 2019.

\bibitem{Moore2016q}
R.~K. Moore, R.~Marxer, and S.~Thill, ``{Vocal interactivity in-and-between
  humans, animals and robots},'' \emph{Frontiers in Robotics and AI}, vol.~3,
  no.~61, 2016.

\bibitem{Moore2017g}
R.~K. Moore, S.~Thill, and R.~Marxer, \emph{{Vocal Interactivity in-and-between
  Humans, Animals and Robots (VIHAR)}}.\hskip 1em plus 0.5em minus 0.4em\relax
  Dagstuhl Seminar 16442, 2017, vol.~6, no.~10.

\end{thebibliography}

\end{document}